# Vision Paper: Designing Graph Neural Networks in Compliance with the European Artificial Intelligence Act

Barbara Hoffmann[1], Jana Vatter[2] and Ruben Mayer[1]

[1]*University of Bayreuth, Universitätsstraße 30, 95447 Bayreuth*
[2]*Technical University of Munich, Arcisstraße 21, 80333 Munich*

**Abstract**

The European Union's Artificial Intelligence Act (AI Act) introduces comprehensive guidelines for the development and oversight of Artificial Intelligence (AI) and Machine Learning (ML) systems, with significant implications for Graph Neural Networks (GNNs). This paper addresses the unique challenges posed by the AI Act for GNNs, which operate on complex graph-structured data. The legislation's requirements for data management, data governance, robustness, human oversight, and privacy necessitate tailored strategies for GNNs. Our study explores the impact of these requirements on GNN training and proposes methods to ensure compliance. We provide an in-depth analysis of bias, robustness, explainability, and privacy in the context of GNNs, highlighting the need for fair sampling strategies and effective interpretability techniques. Our contributions fill the research gap by offering specific guidance for GNNs under the new legislative framework and identifying open questions and future research directions.

**Keywords**
Graph Neural Networks, European Artificial Intelligence Act, Data Management, Data Governance, Robustness, Bias, Privacy, Explainability, Explainable Artificial Intelligence

## 1. Introduction

The European Union has taken a significant step forward in the technological domain with the publication of the European Union Artificial Intelligence Act (AI Act) [1]. This legislation is notable for its provision of guidelines aimed at developing frameworks for Artificial Intelligence (AI) and overseeing Machine Learning (ML) practices. These frameworks possess considerable potential, as they could serve as a blueprint for future endeavors in the domain of AI and ML. The legislation introduces several requirements, for instance data management and data governance, robustness of training data and models and human oversight, which highly impacts Graph Neural Network (GNN) training. Therefore, it is important to know which requirements there are, understand their implications on GNN training and build the model accordingly.

GNNs have unique characteristics that warrant a closer examination. Unlike traditional ML models, GNNs operate on graph-structured data, which introduces complexities in data connectivity and relationships. These special characteristics justify the need for a detailed analysis of the AI Act's impact on GNNs. While there are initial studies examining the AI Act's effects on ML systems [2, 3, 4], no specific research has been conducted on GNNs. In this paper, we argue that GNNs pose specific challenges in terms of core requirements of the AI Act, such as data governance, robustness, explainability and privacy, that demand a closer investigation. This gap underscores the importance of our work in providing tailored guidance for GNNs under the new legislative framework and discussing open issues that demand further research.

In detail, our contributions are:

- Detailed examination of the requirements of the AI Act, tailored precisely to the use of GNNs to ensure compliance with the AI Act. This enhances the current discussion of the AI Act to the specifics of GNNs.

- Investigation of data and model bias in graphs and GNNs, particularly due to unfair sampling during GNN training. This opens a new perspective on data management techniques for GNNs beyond model accuracy and training runtime.

- Exploration of human oversight and explainability with concrete examples of GNN decision-making. We highlight the inherent trade-off between the comprehensibility and accuracy of an explanation, and articulate the demand for further (user) studies to better understand the effect of explanations on AI system stakeholders.

- Analysis of privacy-preserving techniques designed for GNN training. In particular, we stress that while adding noise to features and labels via Differential Privacy techniques is well-explored,

- GNNs additionally expose connections between entities—which can themselves contain privacy-sensitive information.
- Asking important questions arising from our investigations and identifying open research areas in this field. This promises to spawn further research in this area.

In Section 2, we provide an introduction to GNNs and explain the basics of the AI Act, detailing its implications for GNNs. In Section 3, we describe the experiments we conducted and present the results. The open questions that arise from these results are discussed in Section 4. Finally, we review related work in Section 5 and draw conclusions in Section 6.

## 2. Background

This section gives a short introduction to GNNs as well as an overview of the four risk categories delineated by the AI Act, along with their implications for the training of GNNs. Additionally, an overview of the prerequisites outlined in the AI Act for high-risk systems is provided.

### 2.1. Graph Neural Networks

GNNs are specialized neural networks devised for analyzing data represented in graph forms [5, 6], such as networks found in social or citation systems. GNNs process data by transforming the initial node features into embeddings via an iterative mechanism known as message passing. During this process, each node computes new embeddings by aggregating and synthesizing the embeddings from its adjacent nodes, an operation underpinned by neural networks that are trainable at each layer of the GNN. The input features at the initial layer are raw node features, which are systematically enhanced through successive layers to refine the embeddings. This refinement process is driven by the objective of minimizing a predefined loss function, typically optimized through algorithms like stochastic gradient descent. The resulting embeddings, which encapsulate the essential structural and feature-based information of the nodes or the entire graph, are subsequently utilized in downstream applications such as node classification, link prediction, and graph classification. This makes GNNs particularly effective for tasks where data is inherently structured as graphs, such as social networks or citation networks. For a deeper technical explanation of GNNs, Vatter et al. [7] can be consulted.

### 2.2. AI Act Basics

The European Artificial Intelligence Act contains a classification schema for Artificial Intelligence systems. This schema is predicated on a risk-based approach and segregates AI systems into four distinct categories.

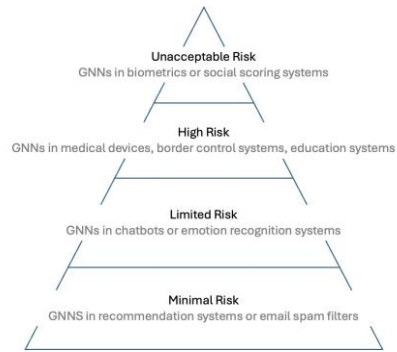

**Figure 1:** Risk Pyramid as described in the AI Act

(1) Minimal Risk: AI systems falling under this category are characterized by their low potential to inflict harm upon the user. Examples of such systems include GNNs utilized in decision support systems, such as recommendation systems [8, 9], or GNNs employed in spam filters for emails [10]. It is noteworthy that the regulations stipulated by the EU AI Act do not extend to this category of AI systems, thereby obviating the need for any regulatory action.

(2) Limited risk: This category refers to risks associated with a lack of transparency in AI usage. AI systems that fall into this category are subject to the requirement of extended transparency. The category comprises for example AI systems that use GNNs in chatbots [11] or emotion recognition systems [12]. The user interacting with these kinds of AI systems needs to be made aware of interacting with a machine so that they can take an informed decision to continue or step back. The content generated by such systems must also be marked as AI generated.

(3) High risk: Encompasses AI systems that process critical personal data or could put the life and health of citizens at risk. Examples of such systems include medical devices [13, 14], systems that control access to education [15], and border control systems. GNNs can be used in all the areas mentioned. For these systems, an enhanced level of quality management is mandated. The criteria for this include data governance, a possibility for human oversight and the robustness of the applications.

(4) Unacceptable Risk: AI systems that fall into this category are deemed illegal. These are for example systems that manipulate the user or contribute to social injustice.

With Article 53, the AI Act also includes a category for General Purpose AI (GPAI). An AI system is classified as GPAI if it is designed to perform a wide variety of tasks across different contexts and applications. This encompasses capabilities like image and speech recognition, audio and video generation, pattern detection, question answering, and translation. GPAI systems are subject to additional documentation and risk management requirements, beyond the standard risk-based requirements. Furthermore, outlined in Article 51, there is a subcategory for Systemic Risk, which is determined by the computing power used during training.

## 2.3. AI Act Requirements for GNNs

The risk category denoted as "High Risk" warrants special attention in this context. The subsequent section provides an overview of the pivotal requirements delineated within the AI Act concerning the training and operation of high-risk GNNs. All of these criteria are outlined in the legal document [1], with a summary provided in Table 1 for quick reference.

| Risk category | Article | Section |
|---|---|---|
| Data Governance | 10 | 2.3.1 |
| Robustness | 15 | 2.3.2 |
| Human Oversight | 14 | 2.3.3 |
| Privacy | (69), (27), (57), 5 | 2.3.4 |

**Table 1**
Overview: Requirements in the AI Act

### 2.3.1. Data Management and Data Governance

The AI Act mandates that the data used in AI systems must comply with data governance requirements. This encompasses data sets utilized for training, validation, and testing purposes.

There is no universally applicable definition of the term *data governance*, neither in the scientific community nor among practitioners in the field of information systems [16, 17]. The definition provided by the AI Act involves ensuring that the data used for training, validation, and testing is relevant, representative, accurate, and as error-free as possible. This includes practices such as proper data collection and preparation, detecting and mitigating bias, and ensuring the data is suitable for the AI system's intended purpose.

In the context of the AI Act, Article 10 delineates the protocols for ensuring effective data governance. It necessitates a comprehensive scrutiny of AI systems, particularly concerning the potential for bias. The manifestation of bias, whether during the training phase or the deployment of AI systems, can precipitate detrimental effects. The primary objective is to forestall any bias that could compromise individual safety and health, infringe upon fundamental rights, or engender discrimination. As per the stipulations of the EU AI Act, such bias must be detected, prevented and mitigated. This is an essential prerequisite for maintaining the integrity and fairness of AI systems.

### 2.3.2. Robustness

According to the AI Act, Article 15, AI systems designed to continue learning post-deployment must be designed to reduce or eliminate the risk of biased outcomes. This requirement also includes ensuring that any potential biases are effectively addressed through suitable risk mitigation strategies. This process of ongoing learning and potential bias reinforcement is referred to as a *feedback loop*.

The robustness of a model refers to its ability to provide consistent and reliable predictions across different data sets and under different conditions. A robust model should also be able to respond well to new, unknown data or to data with minor disturbances. If a model is biased, it is less robust. Bias occurs when a model systematically prefers or excludes specific elements of the data, which may arise from imbalances in training data, errors in model design, or suboptimal sampling techniques.

### 2.3.3. Explainability for Humans

Article 14 of the AI Act stipulates that AI systems must be engineered to enable effective human oversight, ensuring the minimization of risks to health, safety, and fundamental rights. The legislation further mandates that the outputs of these AI systems should be interpretable and the derivation of the results should be understandable. Many AI models, in particular deep neural networks, are referred to as a black box. To make decisions comprehensible, existing interpretability techniques and methodologies from the domain of explainable artificial intelligence (xAI) can be utilized.

Common xAI methods include model-agnostic techniques like Local Interpretable Model Agnostic Explanations (LIME) [18], which approximates black-box models locally with interpretable ones, and SHapley Additive exPlanations (SHAP) [19], which assigns importance values to features based on cooperative game theory. Model-specific methods include feature importance for tree-based models [20] and visualization techniques like saliency maps for neural networks [21], all aimed at increasing the interpretability of AI models.

To enhance the interpretability of GNNs, methods like GNNExplainer [22] have been developed. GNNExplainer has been implemented in PyTorch Geometric [23] as well as in the Deep Graph Library (DGL) [24]. This method operates post-hoc, meaning it explains GNN decisions after they are made. It accomplishes this by analyzing subgraphs within the larger input graph. The output consists of a concise subgraph from the original input, along with a selection of node features deemed most influential in driving the model's predictions. These explanations are localized to individual instances, necessitating retraining for each new instance [22].

**Table 2**
Overview of the datasets

| Name | #Nodes | #Edges | Sensitive Attribute | Feature Size | Class Imbal. | Label Imbal. | Label |
|---|---|---|---|---|---|---|---|
| german | 1,000 | 44,484 | gender | 27 | -0.380 | -0.298 | high/low credit risk |
| recidivism | 18,876 | 642,616 | ethnos | 18 | 0.013 | -0.033 | bail/no bail |
| credit | 30,000 | 304,754 | age | 13 | -0.821 | -0.648 | payment default/no default |
| pokec-n | 66,569 | 1,100,663 | region | 266 | -0.422 | -0.021 | working field |
| pokec-z | 67,796 | 1,303,712 | region | 277 | -0.297 | -0.021 | working field |

### 2.3.4. Privacy

The EU AI Act contains various passages for the secure handling of personal data, emphasizing its protection and confidentiality. For instance, legal regulations mandate enhanced safeguarding of data utilized in the creation of AI systems or in the mitigation of bias within these systems. In the context of AI systems, adherence to all relevant data protection laws, such as the General Data Protection Regulation (GDPR), is obligatory. This safeguarding can be accomplished through the implementation of anonymization and encryption techniques, which in GNNs can be done by the anonymization of the features or by adding noise to the graph, for example by adding or removing edges or nodes [6, 25].

## 3. Analysis

In this section, we examine the areas of influence — Data Governance, Robustness, Explainability, and Privacy — identified in Section 2.3 with regard to their precise impact on GNNs. The focus hereby lies on bias in the data and the model, human oversight and privacy.

### 3.1. Data Governance

High-risk AI systems must be trained with data that meets certain standards and requirements. An important point here is an investigation into possible biases that could affect the health and safety of individuals, have a negative impact on fundamental rights or lead to discrimination prohibited by the AI Act, especially if the data outputs influence the inputs for future operations [1].

If training data exhibits an unbalanced feature or label distribution, this can lead to the aforementioned bias. Technically speaking, any dataset that shows an unequal distribution among its classes can be regarded as imbalanced. Yet, the prevalent view within the community is that imbalanced data specifically refers to datasets with substantial disparities between classes. This specific type of inequality is known as between-class imbalance. Another form of imbalance is the within-class imbalance, which focuses on the distribution of representative data for various subconcepts within a single class [26]. In addition to the class imbalance there is also a label imbalance. The Difference in Proportions of Labels (DPL) metric compares the proportion of observed outcomes with positive labels in one subgroup to the proportion in another subgroup within a training dataset [27].

In this paper we focus on between-class imbalance as well as label imbalance and measure these values according to existing implementations [27, 28]. The results are shown in Table 2. The following applies to both imbalance metrics: the closer to zero, the better the distribution of the data set; the further away from zero, the greater the imbalance. As can be seen in Table 2, some datasets yield low class and label imbalance, while others show high imbalance in classes, labels, or both.

**Key Takeaway:** Class and label imbalance are common issues across various graph datasets. It is important to monitor such imbalance, and, if appropriate, take corrective action.

### 3.2. Robustness

Sampling is a method to efficiently train GNNs on large-scale graphs. When performing sampling during GNN training, a subset of data points - such as nodes, edges, or subgraphs - is selected from the full graph. Before each training epoch, new samples are constructed. As numerous sampling strategies with different objectives exist, the choice of sampling method can significantly influence both the performance and the bias of the model. Inadequate or disproportionate sampling may result in the neglect of crucial segments of the graph, thereby distorting the overall representation of the data. This, in turn, affects the model's ability to generalize and perform accurately, underscoring the intricate link between sampling strategies and the robustness of machine learning models [29]. In the following, we explore whether sampling can influence the bias during training as only a selected subset of nodes and edges is used for training. This could lead to an under-representation of certain classes or labels. For our experiments, we use the datasets german, recidivism, credit, pokec-n and pokec-z as well as the sampling strategies Neighbor [30], VR-GCN [31], LABOR [32], and ShaDow [33]. We additionally include *no sampling* as a baseline. The 2-layer GCN and the sampling strategies are implemented with DGL. Our evaluation is based on the Area Under the Curve (AUC), Statistical Parity [34] and Equality of Opportunity [35]. While Statistical Parity measures how independent the predictions of a model are to a sensitive attribute, Equality of Opportunity denotes to which extent the predictions are performed equally well across all attributes. For both fairness metrics, lower values indicate a fairer model, while for AUC, higher values are better.

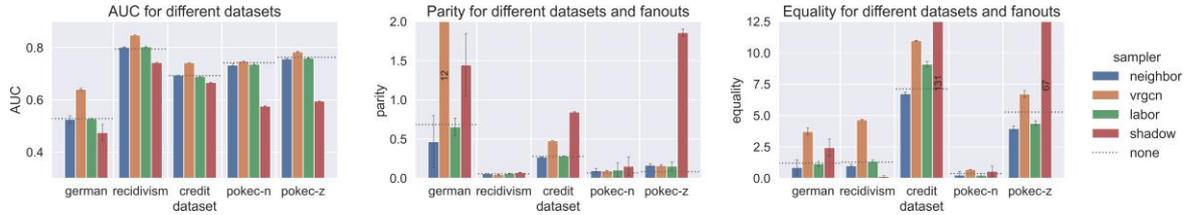

**Figure 2:** Results for the metrics AUC, parity, and equality for the different datasets and samplers. AUC: higher values are better (1.0 is best). Parity and equality: lower values are better (0.0 is best).

In Figure 2, we show the experimental results. Across all datasets and metrics, the values of Neighbor sampling and LABOR usually are close to the baseline (*no sampling*). For parity and equality, they sometimes even lead to better results than the baseline, with the exception of LABOR leading to worse equality on credit. VR-GCN, on the other hand, has a higher AUC score than the baseline and other strategies, but can result in higher values of parity and equality, especially when using the german or credit credit graph. The fourth sampling strategy, namely ShaDow, proves less suitable. A larger bias is induced compared to the other methods, especially for the german, credit, and pokec-z dataset, while the AUC is lower than the baseline.

Neighbor sampling chooses the nodes and edges at random which is beneficial for the model bias since all groups and attributes are treated equally by the sampling method. VR-GCN also is a node-wise method, but with importance scores. As VR-GCN is based on historical activations, valuable information is preserved during the sampling step, but bias can be reinforced. LABOR is a layer-wise strategy using a specialized optimization method and restricts the size of the neighborhood to a small number. Therefore, a higher AUC can be achieved, but the model might not be as fair as with other methods due to the specialized selection of nodes and edges. The fourth method, namely ShaDow, works in a subgraph-based fashion and aims to sample shallow subgraphs with a depth typically around 2 or 3. This could lead to a loss of information needed for training and higher parity and equality values.

**Key Takeaway:** Our experiments have shown that bias can be induced or reinforced when using sampling-based GNN training. Some strategies lead to higher performance values, but also to a more biased model. Robustness against model bias needs to be taken into account when designing GNN sampling methods. This aspect has often been neglected.

### 3.3. Explainability

In this section, we delve deeper into the possibilities for GNN explanation. Our exploration is based on the two implementations of GNNExplainer outlined in Section 2.3.3 which allow for the visualization of subgraphs and feature importance.

GNNExplainer is capable of being applied to various scenarios, including graph classification and link prediction. In this paper, we restrict the scope of our experiments to node classification. For the explanations provided, a basic Graph Convolutional Network (GCN) consisting of two convolutional layers was utilized. This network integrates linear transformations with Rectified Linear Unit (ReLU) activation functions and incorporates dropout to enhance generalization. We used the dataset german as the foundational data for these experiments. The dataset offers insights into whether a customer with specific features qualifies as a good customer with low credit risk. Similarly, the node classification addresses the same question: determining whether the selected node, representing a customer, is credit-worthy or not.

The experiments were conducted using both 2-hop and 1-hop neighborhoods in PyTorch Geometric[1]. Figure 3 illustrates the explanations generated from these experiments. It is evident that increasing the number of hops results in a more complex explanatory graph. Figure 3b shows the 2-hop result in a representation that is essentially imperceptible to the human eye, potentially making it challenging for many stakeholders of an AI system to comprehend. For the 1-hop neighborhood (Figure 3a), the graph remains simple and comprehensible for human interpretation; however, it raises the question of whether such simplicity adequately captures the complexity of node classification. In both scenarios, only the nodes that impacted the decision are depicted, and it remains uncertain whether this is sufficient for a meaningful explanation. For further analysis, nodes can be mapped to their corresponding features, of which an excerpt is shown in Table 4.

---
[1] The visualizations generated by DGL are consistent in content, despite differences in their layout. Due to this uniformity in content, they are not depicted in the paper.

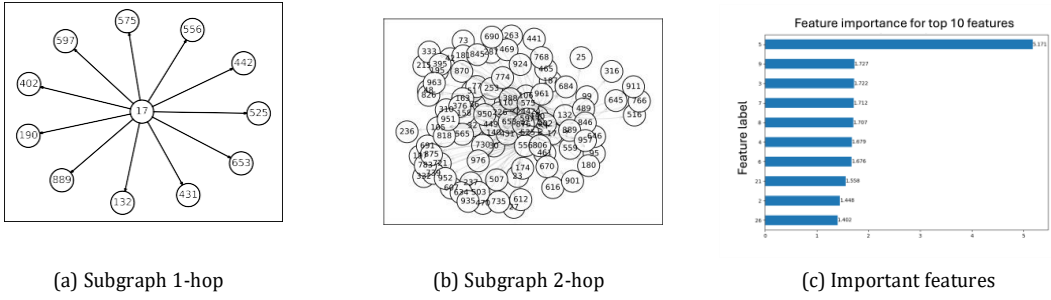

(a) Subgraph 1-hop  (b) Subgraph 2-hop  (c) Important features

**Figure 3:** Overview of PyTorch Geometric generated grahps and important features

The features of the nodes are pivotal in the process of node classification. The visualization output of PyTorch Geometric is displayed in Figure 3c, which shows the most important features leading to a specific node's classification[2]. Additionally, the mapping of feature numbers to their corresponding meanings is detailed in Table 3.

**Key Takeaway:** Graphs are inherently complex structures. Consequently, methods that elucidate the behavior of a GNN through a graph also tend to be intricate. Simplified explanations, such as using only a 1-hop neighborhood or focusing solely on feature importance without including graph information, are more straightforward to comprehend but provide less detailed information. So there is an inherent trade-off between simplicity and the depth of information in GNN explanations.

| Feature Number | Meaning |
|---|---|
| 5 | Loan Duration |
| 3 | Single |
| 6 | Purpose of Loan |
| 4 | Age |
| 9 | Years at current home |
| 7 | Loan Amount |
| 8 | Loan Rate as percent of income |
| 21 | Other Loans at store |
| 2 | Foreign worker |
| 26 | Unemployed |

**Table 3**
Mapping of important features

### 3.4. Privacy

When it comes to maintaining privacy within graph-based data, one of the main concerns are the node features. Features must be kept confidential and anonymized using appropriate methods as needed. Before training a GNN, data anonymization techniques can be employed to inhibit any potential identification of individual users, thus protecting user privacy throughout the model training process. When GNNs are employed, they process edges using machine learning. This raises privacy concerns regarding the edges: Should they be considered private data associated with a specific node, or are they exempt from privacy considerations?

Assuming the data is confidential, Differential Privacy (DP) [37] could be employed as a strategy to protect privacy. The fundamental principle of DP is that when querying a dataset consisting of N individuals, the outcome should be, in probabilistic terms, virtually the same as if the query were run on a similar dataset that has either one fewer or one additional individual. This approach ensures the privacy of each individual with a certain probability. To achieve this level of probabilistic indistinguishability, adequate noise is added to the results of the query, masking individual data points while still providing useful aggregate information [38].

---

[2] An alternative method to explain GNNs is through the GraphLIME framework [36]. GraphLIME utilizes an Hilbert-Schmidt Independence Criterion (HSIC) Lasso model to provide localized, nonlinear explanations for the predictions made by GNNs. These explanations are limited to the K most representative features as the explanation for the prediction of a particular node. The approach integrates both the local structure of the graph and nonlinear dependencies to enhance understanding. Despite employing a distinct approach, GraphLIME solely focuses on visualizing the feature importance and the output looks similar to Figure 3c.

**Table 4**
Listing of important nodes with a selection of their features

|     | Credit Worthy | Gender | Foreign Worker | Single | Age | Loan Duration | PurposeOf Loan | Loan Amount | LoanRateAsPercentOfIncome |
| --- | --- | --- | --- | --- | --- | --- | --- | --- | --- |
| 597 | -1 | Male   | 0 | 1 | 36 | 24 | Business | 4241 | 1 |
| 190 | -1 | Male   | 0 | 1 | 54 | 24 | Business | 4591 | 2 |
| 556 | -1 | Female | 0 | 0 | 28 | 18 | NewCar   | 2278 | 3 |
| 439 | -1 | Female | 0 | 0 | 26 | 12 | Business | 609  | 4 |
| 575 | 1  | Female | 0 | 0 | 24 | 15 | Furniture | 2788 | 2 |
| 132 | 1  | Male   | 0 | 1 | 27 | 15 | Furniture | 2708 | 2 |
| 442 | 1  | Male   | 0 | 1 | 29 | 20 | Other    | 2629 | 2 |
| 889 | 1  | Male   | 0 | 1 | 40 | 28 | UsedCar  | 7824 | 3 |
| 525 | 1  | Male   | 0 | 1 | 30 | 26 | UsedCar  | 7966 | 2 |
| 402 | -1 | Male   | 0 | 1 | 27 | 24 | Business | 8648 | 2 |
| 653 | -1 | Male   | 0 | 1 | 42 | 36 | NewCar   | 8086 | 2 |

When applying DP to GNNs, several challenges arise. As outlined in Section 3.4, DP involves adding noise to the data, which, in the context of graphs, could mean adding or removing edges. Such modifications can significantly alter the dataset. If DP is implemented in a GNN, it is essential to evaluate whether the anonymity provided compromises the utility of the model. In cases of uncertainty, the importance of maintaining privacy versus the utility of the data must be carefully weighed. Moreover, it is crucial to determine the maximum amount of noise that can be introduced before the data loses its meaning due to excessive alteration. This threshold may vary from one GNN to another, as different structures have different tolerances for noise. However, these questions are largely unexplored.

**Key Takeaway:** Privacy in GNNs concerns not only the node features, but also the structural information of the graph itself, i.e., the edges. More research is needed on privacy-preserving GNN training and inference.

## 4. Open Questions and Research Directions

In this section, we look at the questions that arise from our investigations and experiments and propose new research directions.

### 4.1. Bias and Robustness

For our experiments, we use standard sampling methods which do not specifically aim at reducing bias. Our results show that fairness can highly depend on the chosen sampling method. Consequently, the question arises how to better ensure fairness during sampling. Fair random walk strategies, such as those proposed by Rahman et al. [39] and Zhang et al. [40] could be considered in GNN sampling. Further, our experiments are evaluated with metrics commonly used in the field of machine learning. However, when using graphs and GNNs, other factors such as feature distribution and structure, particularly the nature of the connections, play a crucial role in regards to fairness. More research is needed in the directions of designing fairness metrics adapted to the specific characteristics of GNNs.

Furthermore, it is essential to consider whether GNNs are robust against data distribution shifts. A data distribution shift occurs when the data a model uses changes over time, leading to a decline in prediction accuracy. Ensuring robustness in GNNs means maintaining accurate classification performance with new and evolving data.

We summarize the open questions as follows: How can sampling strategies be specifically adapted and optimized for different types of graph data to ensure comprehensive fairness without compromising model performance? What additional or refined fairness metrics need to be developed? What specific adaptations are needed for GNNs to handle dynamic and evolving graph data, and how can these techniques be seamlessly integrated into GNN frameworks?

### 4.2. Explainability and Privacy

Several questions arise in the context of explainability. Firstly, for whom the output of the AI system needs to be explained. Various stakeholders could come into consideration: The end customer, who may be influenced by a system's decision, the employee of the company in which the AI system is used and the auditor of a supervisory authority, who examines compliance with the AI Act in companies, could all be equally interested.

Each of these parties has a unique perspective on the required explanation of an AI decision. Likewise, each of these individuals has their own level of authorization for insights. This implies several privacy issues, as it must be clarified who has the right to view specific data. To illustrate this concept with a specific example: In an online social network, User A receives the explanation that Group G was suggested to him because contact B and his contact C - who is not directly connected to A - are also in similar groups. This explanation allows User A to gain insight into contact C's affiliations, even though C has not directly shared this information with A.

It is also important to evaluate whether the selected explanatory approaches are appropriate. While explanations using subgraphs and key features are commonly employed, alternative forms such as text or images might be more comprehensible to some groups of users. Additionally, the methods we analyzed only highlight the important nodes and features. However, having an overview of the *unimportant* ones might also be beneficial, as it could provide insights for improving the training process of an AI model. To make an informed assessment of which method is preferred and which information in detail would be helpful for different user groups, conducting a user study would be essential [41]. Finally, assessing the effectiveness of explanability methods for human oversight is an interdisciplinary effort that requires further research [42], especially for GNNs which are potentially much more difficult to explain and understand due to their graph structure.

## 5. Related Work

Various papers have already focused on presenting the content of the AI Act in an understandable way [43, 44] or explained the impact of the AI Act on ML systems [2, 3, 4]. The sub-topics we have identified as challenges for GNN training have also been highlighted in the literature. For example, [45] and [46] address bias in GNNs. As mentioned in Section 2.3.3, several methods exist for making GNNs explainable, including GNNExplainer [22] and GraphLiME [36], which we use in our work. Other approaches focus on more theoretical aspects of explanations [47]. Studies have also tackled the security and privacy of GNNs, either by explaining security vulnerabilities and privacy-enhancing measures, as seen in survey articles [48, 49] or practical implementations such as SecGNN [50].

The intersection of the two topics, namely how the requirements set out specifically in the AI Act affect GNNs, has not yet been investigated. We have conducted preliminary analyses of this gap and identified several open questions.

## 6. Conclusion

The AI Act establishes important legal requirements for AI and ML, impacting significant areas within these fields. In our paper, we demonstrated that the AI Act also presents critical challenges for GNNs. Our initial research on this topic has uncovered numerous additional open questions that need to be addressed.